\documentclass[runningheads]{llncs}
\usepackage{graphicx}
\usepackage{multirow}
\begin{document}
\title{Distilling effective supervision for robust medical image segmentation with noisy labels}
\titlerunning{Robust Medical Image Segmentation with PINT}
%
\author{Jialin Shi \and
Ji Wu}

\author{Jialin Shi\inst{1} \and
Ji Wu\inst{1,2}}
\authorrunning{Shi. et al.}
%
\institute{Department of Electronic Engineering, Tsinghua University, Beijing, China\\
\email{shi-jl16@mails.tsinghua.edu.cn}
\and
Institute for Precision Medicine, Tsinghua University, Beijing, China\\
\email{wuji\_ee@mail.tsinghua.edu.cn}}

%
\maketitle              
\begin{abstract}
Despite the success of deep learning methods in medical image segmentation tasks, the human-level performance relies on massive training data with high-quality annotations, which are expensive and time-consuming to collect. The fact is that there exist low-quality annotations with label noise, which leads to suboptimal performance of learned models. Two prominent directions for segmentation learning with noisy labels include pixel-wise noise robust training and image-level noise robust training. In this work, we propose a novel framework to address segmenting with noisy labels by distilling effective supervision information from both pixel and image levels. In particular, we explicitly estimate the uncertainty of every pixel as pixel-wise noise estimation, and propose pixel-wise robust learning by using both the original labels and pseudo labels. Furthermore, we present an image-level robust learning method to accommodate more information as the complements to pixel-level learning. We conduct extensive experiments on both simulated and real-world noisy datasets. The results demonstrate the advantageous performance of our method compared to state-of-the-art baselines for medical image segmentation with noisy labels.

\keywords{3D segmentation \and noisy labels \and robust learning.}
\end{abstract}
\section{Introduction}
Image segmentation plays an important role in biomedical image analysis. With rapid advances in deep learning, many models based on deep neural networks (DNNs) have achieved promising segmentation performance \cite{ref1}. The success relies on massive training data with high-quality manual annotations, which are expensive and time-consuming to collect. Especially for medical images, the annotations heavily rely on expert knowledge. The fact is that there exist low-quality annotations with label noise. Many studies have shown that label noise can significantly affect the accuracy of the learned models \cite{ref2}. In this work, we address the following problem: how to distill more effective information on noisy labeled datasets for the medical segmentation tasks?

Many efforts have been made to improve the robustness of a deep classification model from noisy labels, including loss correction based on label transition matrix \cite{ref3,ref4,ref5}, reweighting samples \cite{ref6,ref7} , selecting small-loss instances \cite{ref8,ref9}, etc. Although effective on image classification tasks, these methods cannot be straightforwardly applied to the segmentation tasks \cite{ref10}. 

There are some deep learning solutions for medical segmentation with noisy labels. Previous works can be categorized into two groups. Firstly, some methods are proposed to against label noise using pixel-wise noise estimation and learning. For example, \cite{ref11} proposed to learn spatially adaptive weight maps and adjusted the contribution of each pixel based on meta-reweighting framework. \cite{ref10} proposed to train three networks simultaneously and each pair of networks selected reliable pixels to guide the third network by extending the co-teaching method. \cite{ref12} employed the idea of disagreement strategy to develop label-noise-robust method, which updated the models only on the pixel-wise predictions of the two models differed. The second group of methods concentrates on image-level noise estimation and learning. For example, \cite{ref13} introduced a label quality evaluation strategy to measure the quality of image-level annotations and then re-weighted the loss to tune the network. To conclude, most existing methods either focus on pixel-wise noise estimation or image-level quality evaluation for medical image segmentation. 

However, when evaluating the label noise degree of a segmentation task, we not only judge whether image-level labels are noisy, but also pay attention to which pixels in the image have pixel-wise noisy labels. There are two types of noise for medical image segmentation tasks: pixel-wise noise and image-level noise. Despite the individual advances in pixel-wise and image-level learning, their connection has been underexplored. In this paper, we propose a novel two-phase framework PINT (Pixel-wise and Image-level Noise Tolerant learning) for medical image segmentation with noisy labels, which distills effective supervision information from both pixel and image levels. 

Concretely, we first propose a novel pixel-wise noise estimation method and corresponding robust learning strategy for the first phase. The intuition is that the predictions under different perturbations for the same input would agree on the relative clean labels. Based on agreement maximization principle, our method relabels the noisy pixels and further explicitly estimates the uncertainty of every pixel as pixel-wise noise estimation. With the guidance of the estimated pixel-wise uncertainty, we propose pixel-wise noise tolerant learning by using both the original pixel-wise labels and generated pseudo labels. Secondly, we propose image-level noise tolerant learning for the second phase. For pixel-wise noise-tolerant learning, the pixels with high uncertainty tends to be noisy. However, there are also some clean pixels which show high uncertainty when they lie in the boundaries. If only pixel-wise robust learning is considered, the network will inevitably neglect these useful pixels. We extend pixel-wise robust learning to image-level robust learning to address this problem. Based on the pixel-wise uncertainty, we calculate the image-level uncertainty as the image-level noise estimation. We design the image-level robust learning strategy according to the original image-level labels and pseudo labels.  Our image-level method could distill more effective information as the complement to pixel-level learning. Last, to show that our method improves the robustness of deep learning on noisy labels, we conduct extensive experiments on simulated and real-world noisy datasets. Experimental results demonstrate the effectiveness of our method.

\begin{figure}[t]
\centering
\includegraphics[scale=0.37]{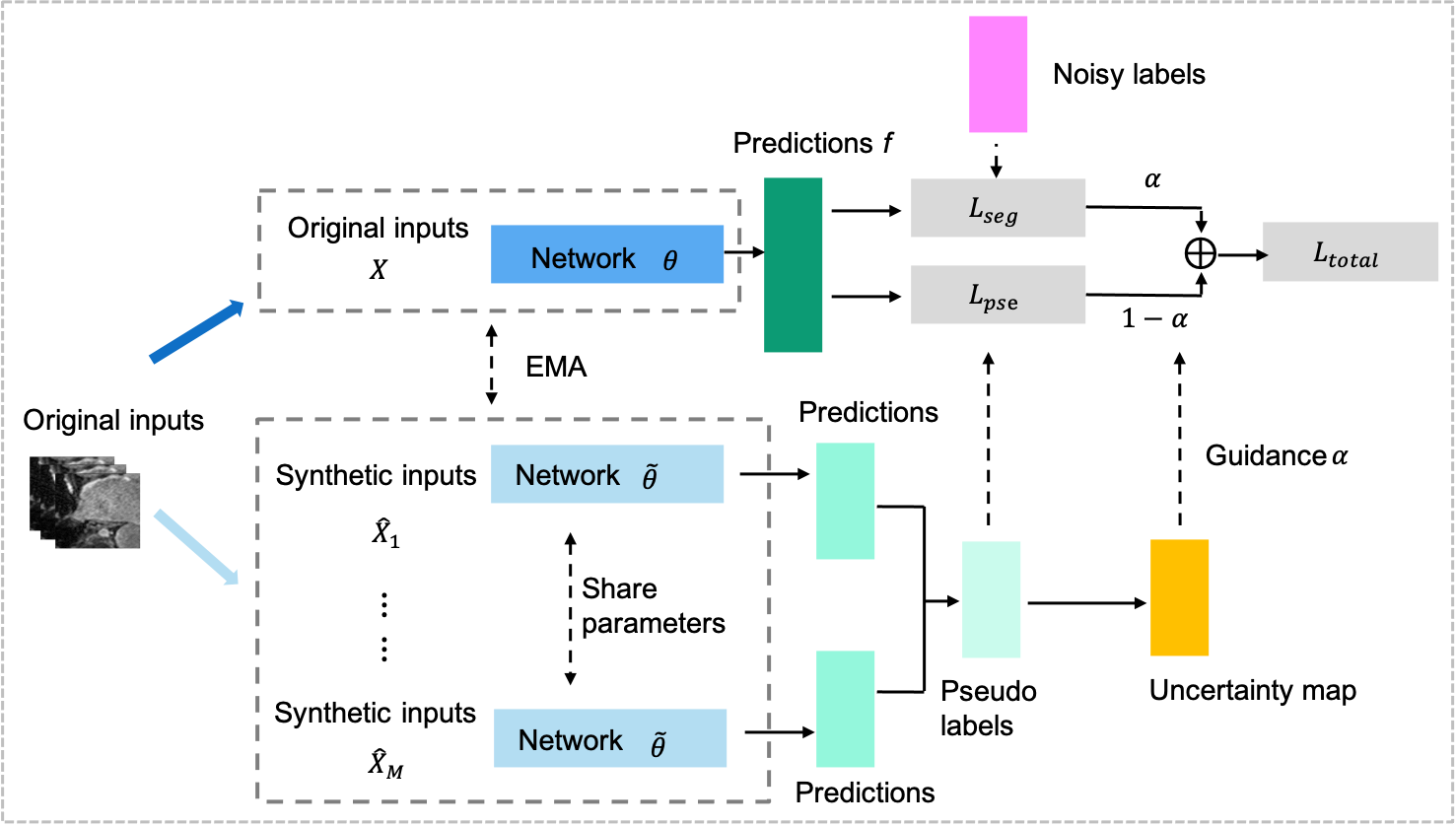}
\caption{Illustration of proposed pixel-wise noise tolerant learning framework. We generate multiple mini-batches of synthetic inputs ${\{\hat{X}_m}\}_{m=1}^M$ with different perturbations. The synthetic mini-batch images go through the network $\widetilde{\theta}$ to get their predictions. 
We regard the mean predictions as the pseudo labels and choose the predictive entropy as the metric to estimate uncertainty. The rectified total loss is calculated with $L_{seg}$ and $L_{pse}$ guided by factor $\alpha$ based on uncertainty map. The image-level noise tolerant learning has the similar pipeline.}


\label{fig1}
\end{figure}

\section{Method}
\subsection{Pixel-wise robust learning}
\textbf{Pixel-wise noise estimation.} 
We study the segmentation tasks with noisy labels for 3D medical images. To satisfy the limitations of GPU memory, we follow the inspiration of mean-teacher model \cite{ref14}. We formulate the proposed PINT approach with two deep neural networks. The main network is parameterized by $\theta$ and the auxiliary network is parameterized by $\widetilde{\theta}$, which is computed as the exponential moving average (EMA) of the $\theta$. At training step $t$, $\widetilde{\theta}$ is updated with $\widetilde{\theta}_t=\gamma\widetilde{\theta}_{t-1}+(1-\gamma)\theta_t $, where $\gamma$ is a smoothing coefficient. Fig.1 shows the pixel-wise noise tolerant learning framework.

For each mini-batch of training data, we generate synthetic inputs ${\{\hat{X}_m}\}_{m=1}^M$ on the same images with different perturbations. Formally, we consider a mini-batch data $(X,Y)$ sampled from the training set, where $X=\{x_1,\cdots,x_K\}$ are $K$ samples, and $Y=\{y_1,\cdots,y_K\}$ are the corresponding noisy labels. In our study, we choose Gaussian noises as the perturbations. Afterwards, we perform $M$ stochastic forward passes on the auxiliary network $\widetilde{\theta}$ and obtain a set of probability vector ${\{p_m\}}_{m=1}^M$ for each pixel in the input. 
In this way, we choose the mean prediction as the pseudo label of $v$-th pixel: $\hat{p}_v=\frac{1}{M}\sum_{m}{p_m^v}$, where $p_m^v$ is the probability of the $m$-th auxiliary network for $v$-th pixel. Inspired by the uncertainty estimation in Bayesian networks \cite{ref15}, we choose the entropy as the metric to estimate the uncertainty. When a pixel-wise label tends to be clean, it is likely to have a peaky prediction probability distribution, which means a small entropy and a small uncertainty. Conversely, if a pixel-wise label tends to be noisy, it is likely to have a flat probability distribution, which means a large entropy and a high uncertainty. As a result, we regard the uncertainty of every pixel as pixel-wise noise estimation:
\begin{equation}
u_v=\mathcal{E}[-\hat{p}_v log \hat{p}_v]
\end{equation}
where $u_v$ is the uncertainty of $v$-th pixel and $\mathcal{E}$ is the expectation operator. The relationship between label noise and uncertainty is verified in Experiments 3.2.

\textbf{Pixel-wise loss.} We propose pixel-wise noise tolerant learning. Considering that the pseudo labels obtained by predictions also contain noisy pixels and the original labels also have useful information, we train our segmentation network leveraging both the original pixel-wise labels and pesudo pixel-wise labels. For the $v$-th pixel, the loss is formulated by:
\begin{equation}
    L_v=\alpha_v L_v^{seg}+(1-\alpha_v)L_v^{pse}
\end{equation}
where $L_v^{seg}$  is the pixel-wise loss between the prediction of main network $f_v$ and original noisy label $y_v$; $L_v^{seg}$ adopts the cross-entropy loss and is formulated by: $    L_v^{seg}=\mathcal{L}_{ce}(f_v,y_v)$ 
$=\mathcal{E}[-y_v log f_v]$. 
$L_v^{pse}$ is the pixel-wise loss between the prediction $f_v$ and pseudo label ${\hat{y}}_v$. ${\hat{y}}_v$ is equal to ${\hat{p}}_v$ for soft label and is the one-hot version of ${\hat{p}}_v$ for hard label. $L_v^{pse}$ is designed as pixel-level mean squared error (MSE) and is formulated by: $ L_v^{pse}=\mathcal{L}_{mse}(f_v,{\hat{y}}_v)=\mathcal{E}[||f_v-{\hat{y}}_v||^2]$. $\alpha_v$ is the weight factor which controls the importance of $L_v^{seg}$  and $L_v^{pse}$.
Instead of manually setting a fixed value, we provide automatic factor $\alpha_v$ based on pixel-wise uncertainty $u_v$. We introduce $\alpha_v$ as $\exp(-u_v)$. If the uncertainty has received one large value, this pixel-wise label is prone to be noisy. This factor $\alpha_v$ tends to zero, which drives the model to neglect original label and focus on the pseudo label. In contrast, when the value of uncertainty is small, this pixel-wise label is likely to be reliable. The factor $\alpha_v$ tends to one and the model will focus on the original label. The rectified pixel-wise total loss could be written as:
\begin{equation}
L_v^{total}=\mathcal{E}[\exp(-u_v)L_v^{seg}+(1-\exp(-u_v))L_v^{pse}]
\end{equation}

\subsection{Image-level robust learning}

\textbf{Image-level noise estimation.} For our 3D volume, we regard every slice-level data as image-level data. Based on the estimated pixel uncertainty, the image-level uncertainty can be summarized as: $U_i=\frac{1}{N_i}\sum_{v}{u_v}$,
where $U_i$ is the uncertainty of $i$-th image ($i$-th slice); $v$ denotes the pixel and $N_i$ denotes the number of pixels in the given image. In this case, the image with small uncertainty tends to provide more information even if some pixels involved have noisy labels. The pipeline is similar to pixel-wise framework and the differences lie in the noise estimation method and corresponding robust total loss construction.

\textbf{Image-level loss.} For image-level robust learning, we train our segmentation network leveraging both the original image-level labels and pseudo image-level labels. For the $i$-th image, the loss is formulated by:
\begin{equation}
    L_i=\alpha_i L_i^{seg}+(1-\alpha_i)L_i^{pse}
\end{equation}
where $L_i^{seg}$ is the image-level cross-entropy loss between the prediction $f_i$ and original noisy label $y_i$; $L_i^{pse}$ is the image-level MSE loss between the prediction $f_i$ and pseudo label ${\hat{y}}_i$; Image-level pseudo label ${\hat{y}}_i$ is composed of pixel-level ${\hat{y}}_v$.
$\alpha_i$ is the automatic weight factor to control the importance of $L_i^{seg}$ and $L_i^{pse}$. Similarity, we provide automatic factor $\alpha_i$ as $\exp(-U_i)$ based on image-level uncertainty $U_i$. The rectified image-level total loss is expressed as: 
\begin{equation}
L_i^{total}=\mathcal{E}[\exp(-U_i)L_i^{seg}+(1-\exp(-U_i))L_i^{pse}]
\end{equation}

Our PINT framework has two phases for training with noisy labels. In the first phase, we apply the pixel-wise noise tolerant learning. Based on the guidance of the estimated pixel-wise uncertainty, we can filter out the unreliable pixels and preserve only the reliable pixels. In this way, we distill effective information for learning. However, for segmentation tasks, there are also some clean pixels have high uncertainty when they lie in the marginal areas. Thus, we adopt the image-level noise tolerant learning for the second phase. Based on the estimated image-level uncertainty, we can learn from the images with relative more information. That is, image-level learning enables us to investigate the easily neglected hard pixels based on the whole images. Image-level robust learning can be regarded as the complement to pixel-level robust learning.

\section{Experiments and Results}
\subsection{ Datasets and implementation details}
\textbf{Datasets.} For synthetic noisy labels, we use the publicly available Left Atrial (LA) Segmentation dataset. We refer the readers to the Challenge \cite{ref20} for more details. LA dataset provides 100 3D MR image scans and  segmentation masks for training and testing. We split the 100 scans into 80 scans for training and 20 scans for testing. We randomly crop $112\times 112 \times 80$ sub-volumes as the inputs. All data are pre-processed by zero-mean and unit-variance intensity normalization. 

For real-world dataset, we have collected CT scans with 30 patients (average 72 slices / patient). The dataset is used to delineate the Clinical Target Volume (CTV) of cervical cancer for radiotherapy. Ground truths are defined as the reference segmentations generated by two radiation oncologists via consensus. Noisy labels are provided by the less experienced operators. 20 patients are randomly selected as training images and the remaining 10 patients are selected as testing images. We resize the images to $256\times 256\times64$ for inputs.

\textbf{Implementation details.} The framework is implemented with PyTorch, using a GTX 1080Ti GPU. We employ V-net \cite{ref16} as the backbone network and add two dropout layers after the L-5 and R-1 stage layers with dropout rate $0.5$ \cite{ref17}. We set the EMA decay $\gamma$ as 0.99 referring to the work \cite{ref14} and set batch size as 4. We use the SGD optimizer to update the network parameters (weight decay=0.0001, momentum=0.9). Gaussian noises are generated from a normal distribution. For the uncertainty estimation, we set $M=4$ for all experiments to balance the uncertainty estimation quality and training efficiency. The effect of hyper-parameters M is shown in supplementary materials. Code will be made publicly available upon acceptance.

For the first phase, we apply the pixel-wise noise tolerant learning for 6000 iterations. At this time, the performance difference between different iterations is small enough in our experiments. The learning rate is initially set to 0.01 and is divided by 10 every 2500 iterations. For the second phase, we apply the image-level noise tolerant learning. When trained on noisy labels, deep models have been verified to first fit the training data with clean labels and then memorize the examples with false labels. Following the promising works\cite{ref18,ref19}, we adopt “high learning rate” and “early-stopping” strategies to prevent the network from memorizing the noisy labels. In our experiments, we set a high learning rate as lr=0.01 and the small number of iterations as 2000. All hyper-parameters are empirically determined based on the validation performance of LA dataset.

\begin{table}[!t]
\centering
\caption{Segmentation performance comparison on simulated LA noisy dataset with varying noise rates (25\%, 50\% and 75\%). The average values ($\pm$std) over 3 repetitions are reported. The arrows indicate which direction is better.}
\scalebox{0.84}{
\begin{tabular}{c|cc|cc|cc}
\hline
\multirow{2}{*}{Method}
&\multicolumn{2}{c|}{25\%} 
&\multicolumn{2}{c|}{50\%}
&\multicolumn{2}{c}{75\%}\\
\cline{2-7}
& Dice(\%)$\uparrow$ &ASD $\downarrow$ & Dice(\%)$\uparrow$ &ASD $\downarrow$ & Dice(\%)$\uparrow$ &ASD $\downarrow$\\
\hline
V-net\cite{ref16} & 86.34$\pm$0.59 &2.72$\pm$0.36 &82.55$\pm$0.26& 3.35$\pm$0.01 &72.76$\pm$1.00 &5.48$\pm$0.06 \\
Reweighting \cite{ref11} & 87.31$\pm$0.28 &2.46$\pm$0.35 &83.24$\pm$0.70& 3.20$\pm$0.17 &73.02$\pm$0.32 &5.30$\pm$0.12  \\
Tri-network \cite{ref10} & 87.92$\pm$0.44 &2.37$\pm$0.27 &84.79$\pm$0.44& 2.83$\pm$0.16 &73.88$\pm$0.46 &5.22$\pm$0.11  \\
Pick-and-learn\cite{ref13} & 88.47$\pm$0.30 &1.92$\pm$0.24 &85.09$\pm$0.56& 2.73$\pm$0.20 &73.30$\pm$0.27 &5.11$\pm$0.08  \\
\hline
PNT  & 88.29$\pm$0.43 &1.82$\pm$0.11 &86.16$\pm$0.69& 2.43$\pm$0.05 &74.92$\pm$0.19 &5.16$\pm$0.01  \\
INT  & 89.24$\pm$0.21 &1.75$\pm$0.21 &85.78$\pm$0.55 & 2.56$\pm$0.12 & 74.42$\pm$0.23 &5.20$\pm$0.08  \\
PINT & \textbf{90.49$\pm$0.39} &\textbf{1.60$\pm$0.06} &\textbf{89.04$\pm$0.71} & \textbf{1.92$\pm$0.17} & \textbf{76.25$\pm$0.44} &\textbf{4.56$\pm$0.18}  \\

\hline
\end{tabular}}
\label{table1}
\end{table}

\subsection{Results}

\textbf{Experiments on LA dataset.} We conduct experiments on  LA dataset with simulated noisy labels. We randomly select 25\%, 50\% and 75\% training samples and further randomly erode/dilate the contours with 5-18 pixels to simulate the non-expert noisy labels. We train our framework with non-expert noisy annotations and evaluate the model by the Dice coefficient score and the average surface distance (ASD [voxel]) between the predictions and the accurate ground truth annotations \cite{ref17}. We compare our PINT framework with multiple baseline frameworks. 1) V-net \cite{ref16}: which uses a cross-entropy loss to directly train the network on the noisy training data; 2) Reweighting framework \cite{ref11}: a pixel-wise noise tolerant strategy based on the meta-reweight framework; 3)Tri-network \cite{ref10}:a pixel-wise noise tolerant method based on tri- network extended by co-teaching method. 4) Pick-and-learn framework \cite{ref13}: an image-level noise tolerant strategy based on image-level quality estimation. We use PNT to represent our PINT framework with only pixel-wise robust learning and INT to represent our PINT framework with only image-level robust learning. Our PINT framework contains two-phase pixel-wise and image-level noise tolerant learning. 

Table 1 illustrates the experimental results on the testing data. For clean-annotated dataset, the V-net has the upper bound of average Dice 91.14\% and average ASD 1.52 voxels. (1) We can observe that as the noise percentage increase (from clean labels to 25\%, 50\% and 75\% noise rate), the segmentation performance of baseline V-net decreases sharply. In this case, the trained model tends to overfit to the label noise. When adopting noise-robust strategy, the segmentation network begins to recover its performance. (2) For pixel-wise noise robust learning, we compare Reweighting method \cite{ref11} and our PNT with only pixel-wise distillation. Our method gains 2.92\% improvement of Dice for 50\% noise rate (83.24\% vs 86.16\%). For image-level noise robust learning, we compare Pick-and-learn \cite{ref13} and our INT with only image-level distillation. Our method achieves 1.12\% average gains of Dice for 75\% noise rate (73.30\% vs 74.42\%). These results verify that our pixel-wise and image-level noise robust learning are effective. (3)We can observe that our PINT outperforms other baselines by a large margin. Moreover, comparing to PNT and INT methods, our PINT with both pixel-wise and image-level learning shows better performance, which verifies that our PINT can distill more effective supervision information.


\begin{figure}[t]
\centering
\includegraphics[scale=0.4]{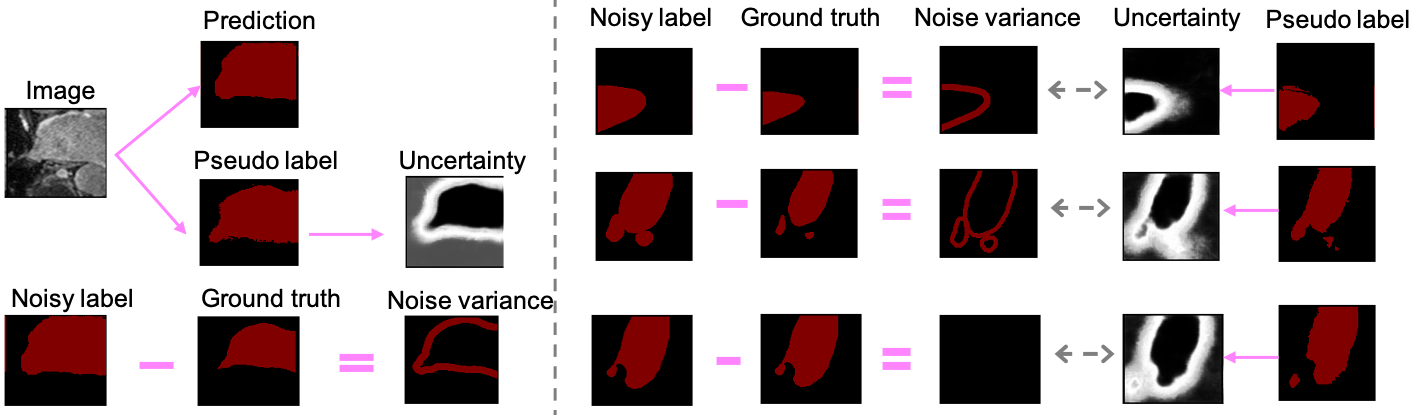}
\caption{Illustration of noise variance and pixel-wise uncertainty. The white color means higher uncertainty. The pixels with high uncertainty usually lie in the noise areas or marginal areas.} 
\label{fig2}
\end{figure}

\textbf{Label noise and uncertainty.} To investigate the relationship between pixel-wise uncertainty estimation and noisy labels, we illustrates the results of randomly selected samples on synthetic noisy LA dataset with 50\% noise rate in Fig.2. 
The discrepancy between ground-truth and noisy label is approximated as the noise variance. We can observe that the noise usually exists in the areas with high uncertainty (shown in white color on the left). Inspired by this, we provide our pixel-wise noise estimation based on pixel-wise uncertainty awareness. Apart from noisy labels, pseudo labels also suffer from the noise effect. The best way for training robust model is to use both original noisy labels and pseudo labels.
Furthermore, multiple examples are shown on the right. We observe that there are some clean pixels show high uncertainty when they lie in the boundaries. If only pixel-wise robust learning is considered, the network will neglect these useful pixels. Therefore, we propose image-level robust learning to learn from the whole images for distilling more effective information.

\textbf{Visualization.} As shown in Fig.3, we provide the qualitative results of the simulated noisy LA segmentation dataset and real-world noisy CTV dataset. For noisy LA segmentation, we show some random selected examples with 50\% noise rate. Compared to the baselines, our PINT with both pixel-wise and image-level robust learning yields more reasonable segmentation predictions. 

\begin{figure}[t]
\centering
\includegraphics[scale=0.4]{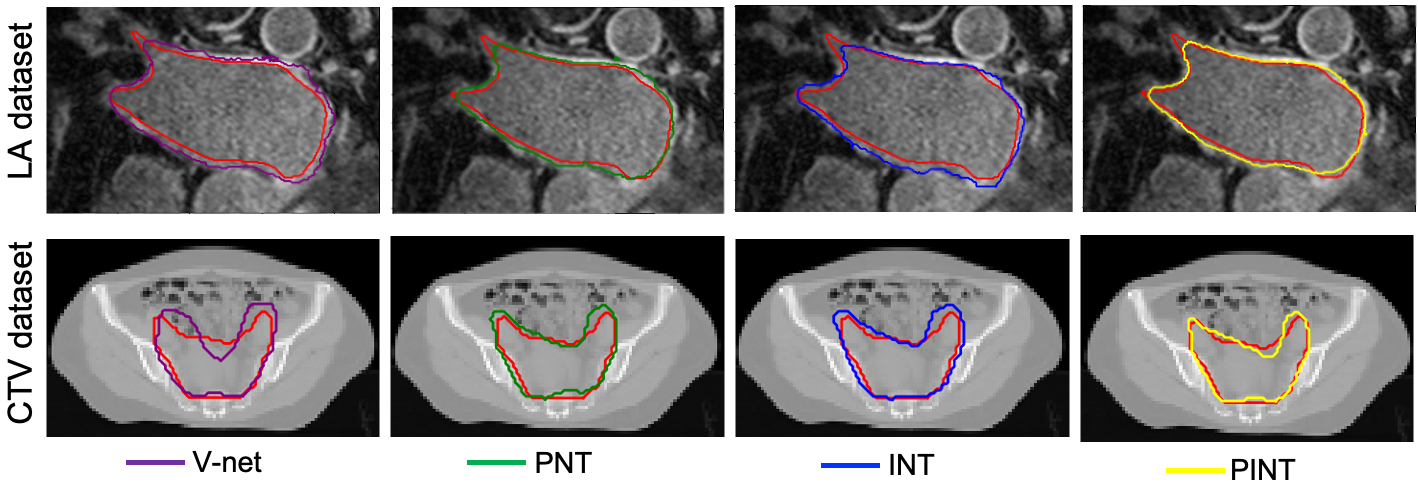}
\caption{Qualitative results of segmentation with noisy labels on simulated LA dataset and real-world CTV dataset. The ground truths, the predictions by V-net, the predictions by PNT, the predictions by INT and the predictions by PINT are colored with red, purple, green, blue and yellow, respectively.} 
\label{fig3}
\end{figure}

\textbf{Experiments on real-world dataset.} We explore the effectiveness of our approach on a real CTV dataset with noisy labels. Due to the lack of professional medical knowledge, the non-expert annotators often generate noisy annotations. The results are shown in Table 2. ‘No noise’ means we train the segmentation network with clean labels. The other methods including V-net, Re-weighting, Pick-and-learn, PNT, INT and PINT are the same with LA segmentation. All the results show that our PINT with both pixel-wise and image-level robust learning can successfully recognize the clinical target volumes in the presence of noisy labels and achieves competitive performance compared to the state-of-the-art methods.

\begin{table}[bh]
\centering
\caption{Segmentation performance comparison on real-world CTV dataset. The arrows indicate which direction is better and the average values ($\pm$std) over 3 repetitions are reported. }
\scalebox{0.75}{
\begin{tabular}{c|cccc|ccc}
\hline
Method &No noise\cite{ref16} &V-net\cite{ref16} &Re-weighting\cite{ref11} &Pick-and-learn\cite{ref13} &PNT &INT &PINT \\
\hline
Dice(\%)$\uparrow$ & 77.26$\pm$0.53 & 68.26$\pm$0.21 &69.31$\pm$0.43 &70.79$\pm$0.31 &73.57$\pm$0.37 &72.08$\pm$0.56 &\textbf{75.31$\pm$0.15} \\
ASD [voxel] $\downarrow$ & 1.38$\pm$0.03  & 2.25$\pm$0.02  &2.05$\pm$0.06 &2.11$\pm$0.07 &1.85$\pm$0.04 &1.92$\pm$0.08 &\textbf{1.76$\pm$0.13}\\
\hline
\end{tabular}}
\label{table2}
\end{table}

\section{Conclusion}
In this paper, we propose a novel framework PINT, which distills effective supervision information from both pixel and image levels for medical image segmentation with noisy labels. We explicitly estimate the uncertainty of every pixel as pixel-wise noise estimation, and propose pixel-wise robust learning by using both the original labels and pseudo labels. Furthermore, we present the image-level robust learning method to accommodate more informative locations as the complements to pixel-level learning. As a result, we achieve the competitive performance on the synthetic noisy dataset and real-world noisy dataset. In the future, we will continue to investigate the joint estimation and learning of pixel and image levels for medical segmentation tasks with noisy labels.

%
%

%
%
%

\end{document}